\documentclass[wcp]{jmlr}
\usepackage{amsmath,amssymb,graphicx,url}
\jmlrvolume{266}
\jmlryear{2025}
\jmlrworkshop{Conformal and Probabilistic Prediction with Applications}
\jmlrproceedings{PMLR}{Proceedings of Machine Learning Research}

\title{When Can We Reuse a Calibration Set for Multiple Conformal Predictions?}
\author{\Name{A.~A. Balinsky}\Email{BalinskyA@cardiff.ac.uk}\\
        \addr{Department of Mathematics, Cardiff University, UK}  \\
\Name{A.~D. Balinsky}\Email{alexander.balinsky22@imperial.ac.uk}\\
	\addr{Department of Computing,
		Imperial College London, UK}
}

\usepackage{float}
\usepackage{graphicx}
\usepackage{subcaption}

\begin{document}
	
	\maketitle

\begin{abstract}  
Reliable uncertainty quantification is crucial for the trustworthiness of machine learning applications. Inductive Conformal Prediction (ICP) offers a distribution-free framework for generating prediction sets or intervals with user-specified confidence. However, standard ICP guarantees are marginal and typically require a fresh calibration set for each new prediction to maintain their validity. This paper addresses this practical limitation by demonstrating how e-conformal prediction, in conjunction with Hoeffding's inequality, can enable the repeated use of a single calibration set with a high probability of preserving the desired coverage. Through a case study on the CIFAR-10 dataset, we train a deep neural network and utilise a calibration set to estimate a Hoeffding correction. This correction allows us to apply a modified Markov's inequality, leading to the construction of prediction sets with quantifiable confidence. Our results illustrate the feasibility of maintaining provable performance in conformal prediction while enhancing its practicality by reducing the need for repeated calibration. The code for this work is publicly available.
\end{abstract}


\section{Introduction}
Reliable uncertainty quantification is paramount in machine learning, and the Inductive Conformal Predictor (ICP) provides a compelling, distribution-free framework for this purpose. Unlike conventional point predictions, ICPs generate prediction sets (for classification) or intervals (for regression) that come with a user-specified confidence level. This is achieved by using a training set to build a base model and a distinct calibration set to estimate the inherent uncertainty. By calculating non-conformity scores, which quantify the "strangeness" of new data relative to the calibration data, ICP constructs prediction regions designed to contain the true outcome with the desired probability, relying only on the mild exchangeability assumption. This inherent reliability makes ICP particularly valuable in applications where the trustworthiness of a prediction is as critical as the prediction itself.

Despite its strengths, the precise nature of the guarantees provided by conformal methods is often misunderstood. As highlighted in \cite{JessicaHullman2024}, a common misconception is that "\textit{conformal prediction provides individualised uncertainty}". It's crucial to emphasise that standard conformal prediction provides only a \textit{marginal coverage guarantee}.

The ICP operates by defining a \textit{non-conformity} measure that assesses the anomaly of a new data point compared to the calibration data. This measure is then used to compute the p-values or e-values for potential outcomes, enabling the construction of prediction sets that satisfy the desired marginal coverage. To be precise, this guarantee holds over the combined set of the calibration data and the new data point, not solely for the new instance. Consequently, repeatedly using the same calibration set for numerous future predictions can compromise this guarantee.

In this article, we aim to demonstrate that by employing the e-conformal predictor, users can, with high probability, retain the same calibration set for multiple future uncertainty quantification tasks while maintaining provable performance.

\section{Background and Methodology}

Conformal Prediction (CP) provides a simple and robust framework for quantifying prediction uncertainty by generating confidence sets or intervals guaranteed to contain the true outcome with a user-specified probability \citep{VovkGammShaf2025}. For readers unfamiliar with the fundamentals of CP, we recommend the recent reviews and introductions found in \cite{10.3150/21-BEJ1447, angelopoulos2023conformal}. Recent advancements in the field include {\em conformal e-prediction} and the {\em BB-predictor} \citep{pmlr-v230-balinsky24a}, which broaden the applicability of conformal methods. Notably, the BB-predictor relaxes the standard exchangeability assumption by operating under the weaker condition of cycle invariance, thereby significantly expanding the range of permissible data-generating processes. In \cite{GauthierBachJordan2025}, the authors highlight several key advantages of conformal e-value prediction (and the BB-predictor) over traditional conformal p-value prediction.

The core objective of CP is to construct conformal sets that, given an unlabeled data point $x \in \mathcal{X}$, contain the true target $y \in \mathcal{Y}$ with a high probability.

Consider a dataset of $n$ exchangeable labeled data points $\{ (x_i, y_i) \}_{i = 1,\ldots,n}$, where each pair $(x_i, y_i)$ is drawn from some distribution over $\mathcal{X} \times \mathcal{Y}$. This dataset, known as the \textit{calibration set}, is used to construct conformal sets. Let $\epsilon \in (0, 1)$ be a predefined error level. Given a new pair $(x_{n+1}, y_{n+1})$, drawn from the same distribution, such that the sequence $(x_1, y_1), \ldots, (x_n, y_n), (x_{n+1}, y_{n+1})$ remains exchangeable, the goal is to construct a conformal set $\hat{C}_n (x_{n+1})$ satisfying:
\begin{equation}\label{CS}
    P(y_{n+1} \in \hat{C}_n (x_{n+1})) \geq 1- \epsilon,
\end{equation}
where the probability is taken over \textbf{both} the calibration set $\{ (x_i, y_i) \}_{i = 1,\ldots,n}$ and the new data point $(x_{n+1}, y_{n+1})$.

A primary source of misconceptions regarding CP arises from the fact that many authors present formula (\ref{CS}) without explicitly defining the probability space. Our true objective is to achieve a guarantee similar to (\ref{CS}) where the probability is conditioned solely on the new data point $(x_{n+1}, y_{n+1})$!

\begin{definition}[Non-conformity Measure]
    A non-conformity measure $L(x,y)$ quantifies the "strangeness" of a labeled data point $(x,y)$ with respect to a trained model. It essentially measures the discrepancy between the model's prediction and the true label.
\end{definition}

In classification tasks, a common non-conformity measure is the negative predicted probability of the true class: $L(x, y) = 1 - \hat{p}(y|x)$, where $\hat{p}(y|x)$ represents the model's predicted probability for the true class $y$ given the input $x$.

Our approach will leverage \textit{e-test statistics}. The fundamental idea behind e-test statistics is a direct application of Markov's inequality: if $Z$ is a non-negative random variable with expectation $\mathbf{E}[Z]$, then for any $\alpha > 0$,
\[
P(Z \geq 1/\alpha) \leq \mathbf{E}[Z] \cdot \alpha.
\]
Letting $\tilde{\alpha} = \mathbf{E}[Z] \cdot \alpha$, we can rewrite this inequality as:
\begin{equation} \label{Markov}
    P\left(Z \geq \frac{\mathbf{E}[Z]}{ \tilde{\alpha}}\right) \leq \tilde{\alpha}.
\end{equation}

We will apply the inequality (\ref{Markov}) to a non-conformity measure $L: \mathcal{X} \times \mathcal{Y} \rightarrow \mathbb{R}_+$. Notice that inequality (\ref{Markov}) involves the expectation of some random variable related to the non-conformity scores, which we cannot compute directly from a finite calibration set. Instead, we can estimate this expectation using the empirical mean calculated from the calibration set. Suppose this estimation of the empirical mean is greater than or equal to the true expectation. In that case, we can potentially reuse the calibration set for future uncertainty predictions without compromising the coverage guarantee. This condition, where the estimation of the empirical mean overestimates the true mean, holds with high probability due to concentration inequalities such as Hoeffding's inequality.

Hoeffding's inequality \citep{hoeffding1963probability} is a cornerstone result in learning theory, providing powerful bounds on the probability that the sum of independent, bounded random variables deviates significantly from its expected value.

\begin{theorem}[Hoeffding's inequality] Let $Z_1, \ldots , Z_n$ be independent bounded
random variables with $Z_i \in [a, b]$ for all $i$, where $-\infty < a \leq b < \infty$.
Then
$$ 
P \left(  \frac{1}{n} \sum_{i=1}^n (Z_i - \mathbf{E}[Z_i]) \geq t \right) \leq \exp \left(  
-\frac{2nt^2}{(b-a)^2}\right)
$$
and
$$
P \left(  \frac{1}{n} \sum_{i=1}^n (Z_i - \mathbf{E}[Z_i]) \leq - t \right) \leq \exp \left(  
-\frac{2nt^2}{(b-a)^2}\right)
$$
for all $t \geq 0$.
\end{theorem}

For our purposes, we will specifically utilize the second inequality to establish an upper bound on the true mean of a random variable using its empirical mean. Recall that Markov's inequality requires an upper bound on the expectation.

Consider a sequence of $n$ independent and identically distributed (i.i.d.) random variables $Z_1, \ldots , Z_n$ where each  $Z_i \in [a, b]$ for all $i$, $-\infty < a \leq b < \infty$, and  has a mean $\mu = \mathbf{E}(Z_i)$.

Applying the second part of Hoeffding's inequality, we have that with probability at least
$1- \exp \left(  
-\frac{2nt^2}{(b-a)^2}\right)$, the following holds:
$$
 \frac{1}{n} \sum_{i=1}^n (Z_i - \mu) \geq -t,
$$
which can be rearranged to give an upper bound on the true mean 
$\mu$:
$$
\mu \leq  \left(\frac{1}{n} \sum_{i=1}^n Z_i\right) +t.
$$

Now, let $Z$ be a non-negative random variable with expectation 
$\mu$. Using the upper bound derived from Hoeffding's inequality, we can modify Markov's inequality (\ref{Markov}). With probability at least $1- \exp \left(  
-\frac{2nt^2}{(b-a)^2}\right)$:
\begin{equation} \label{main-ineq}
P\left(Z \geq \frac{ \left(\frac{1}{n} \sum_{i=1}^n Z_i\right) +t}{ \tilde{\alpha}}\right)\leq    \tilde{\alpha},
\end{equation}

\section{Experiments and Results: A Case Study on CIFAR-10}

The CIFAR-10 dataset \cite{krizhevsky2009learning} is a widely used collection of 60,000 small 32×32 color images categorized into 10 mutually exclusive classes: airplane, automobile, bird, cat, deer, dog, frog, horse, ship, and truck. Each class contains 6,000 images, with 5,000 images allocated for training and 1,000 for testing. This balanced dataset is a popular benchmark for image classification tasks in machine learning and computer vision. 
Let's examine some examples:

\begin{figure}[H]
	\centering
	\includegraphics[width=0.7\linewidth]{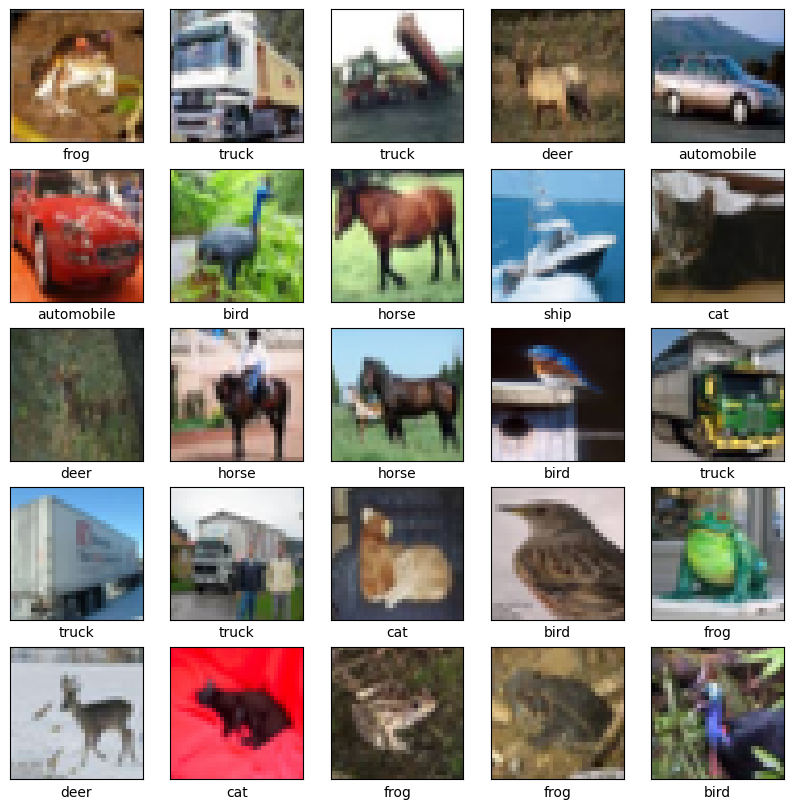}
	\caption{\label{fig:samples} The first twenty five images from the CIFAR-10 database images.}
\end{figure}

\subsection{Simple Deep Neural Network model}

\begin{figure}[H]
	\centering
	\includegraphics[width=1.0\linewidth]{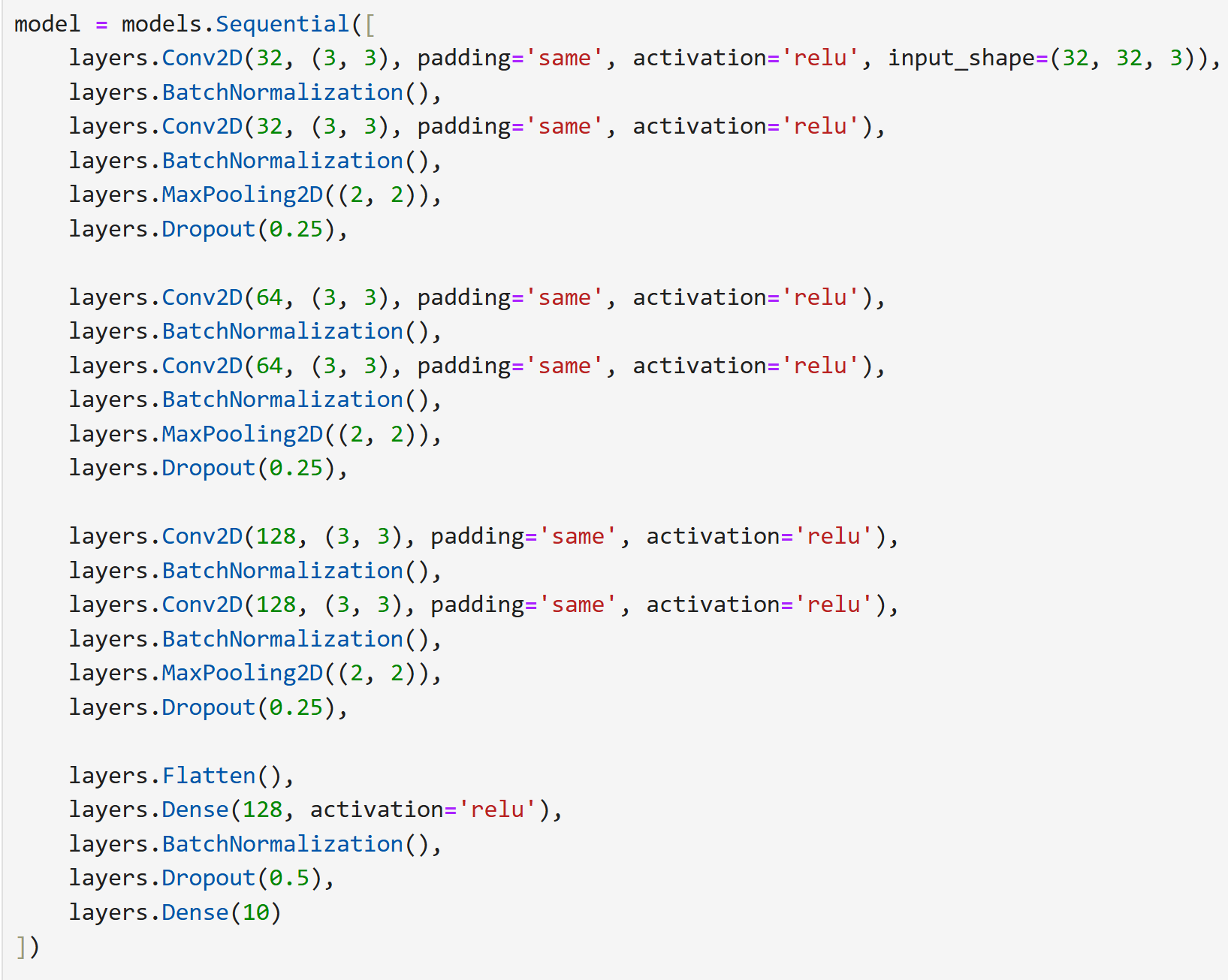}
	\caption{\label{fig:model } Our model.}
\end{figure}

We will train a simple TensorFlow  Keras model  (Figure~\ref{fig:model }) with \textit{SparseCategoricalCrossentropy} loss function and '\textit{adam}' optimizer. 

For reproducibility we use  $tf.random.set\_seed(2025)$.
This model has $552,874$ total number of parameters with
 $551,722$ trainable and 
 $1,152$ not-trainable parameters.

\begin{figure}[H]
	\centering
	\includegraphics[width=0.7\linewidth]{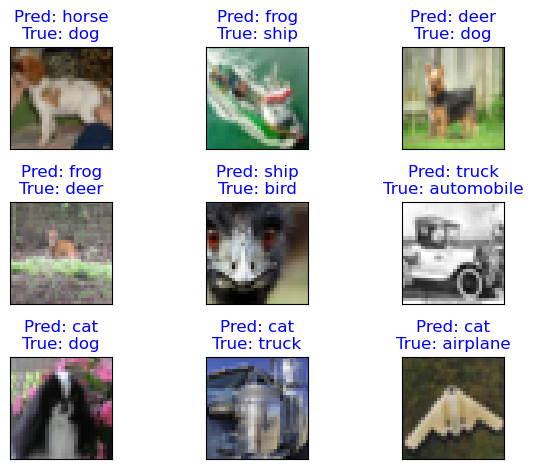}
	\caption{\label{fig:wrong} Some images that the model predicts the wrong labels.}
\end{figure}

Upon training the model with the training images and labels over $50$ epochs, we achieved an accuracy of $94$\% on the training data and $86$\% on the test data. Thus, the model appears to be quite effective. 
Figure~\ref{fig:wrong} shows some instances where the predictions were incorrect (with true and predicted labels).

\subsection{Inductive Conformal Prediction with e-test statistics}

With the model trained and the non-conformity measure defined as the negative predicted probability of the true class, $L(x, y) = 1 - \hat{p}(y|x)$, the focus shifts to the test data.

\begin{figure}[H]
	\centering
	\includegraphics[width=0.45\linewidth]{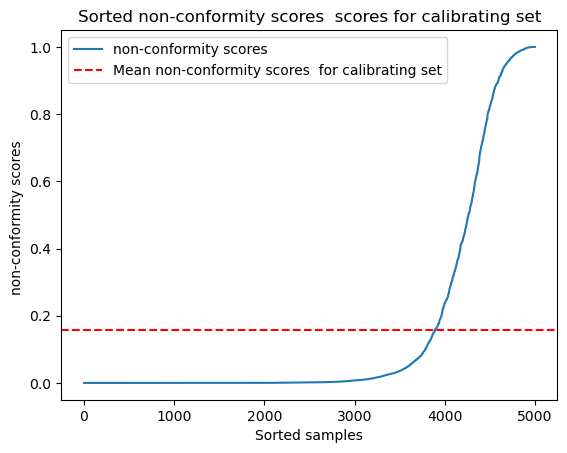}
    \includegraphics[width=0.45\linewidth]{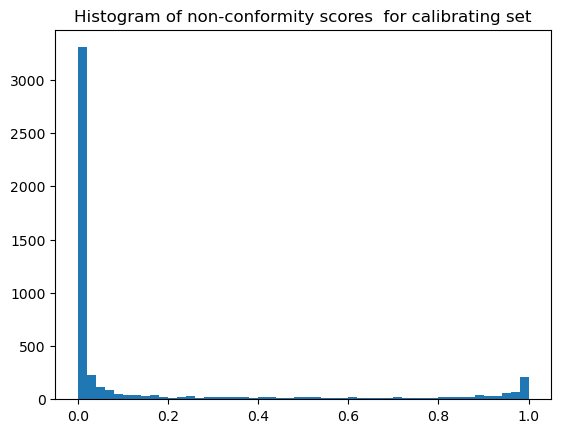}
	\caption{\label{fig:calibr_values} Values of the non-conformity scores and their histogram on the {\em CalibrationSet}.}
\end{figure}

This data is randomly split into two equal subsets: the {\em CalibrationSet} and the {\em ConformalPredictionTestSet}. Figure~\ref{fig:calibr_values} illustrates the sorted non-conformity scores for the {\em CalibrationSet} and their histogram.

The empirical mean value of the non-conformity measure is 
$0.15795493$.

The subsequent step involves estimating the Hoeffding correction $t$ required for inequality (\ref{main-ineq}). Since the non-conformity scores are bounded within the interval $[0, 1]$, the term $(b-a)$ in Hoeffding's inequality is equal to 1. With a calibration set size of $n = 5000$, the probability that inequality (\ref{main-ineq}) holds is given by $1 - \exp\left(-\frac{2nt^2}{(b-a)^2}\right) = 1 - \exp(-2 \cdot 5000 \cdot t^2) = 1 - \exp(-10000 t^2)$.

By choosing $t = 1/50$, we obtain a probability of $1 - \exp(-10000 \cdot (1/50)^2) = 1 - \exp(-4) \approx 0.98$. This suggests that we can be approximately 98\% confident in using this calibration set for all new data points with a Hoeffding correction of $t = 1/50$ in inequality (\ref{main-ineq}).

With the empirical mean of $0.15795493$ and $t = 1/50$,
the numerator in (\ref{main-ineq}) is equal $\approx 0.178$
If we select $\tilde{\alpha} = 0.2$, the inequality (\ref{main-ineq}) implies that
\begin{equation} \label{main_cr}
    P\left(L \leq 0.89 \right) >   0.8.
\end{equation}
We construct prediction sets accordingly by applying this threshold across the entire {\em ConformalPredictionTestSet}.
\begin{table}[H]
\centering
\caption{Label set size distribution across the entire {\em ConformalPredictionTestSet}.}

\begin{tabular}{ |l|c|c|c|c|c|c|c|c|c|c|c| } 
 \hline
 Size of label set & 0 & 1 & 2 & 3 & 4 & 5 & $>$ 5 \\ 
 \hline
Number of examples & 0 & 4150 & 663 & 160 & 26 & 1 & 0 \\
 \hline
\end{tabular}
\end{table}

\begin{figure}[H]
	\centering
	\includegraphics[width=0.4\linewidth]{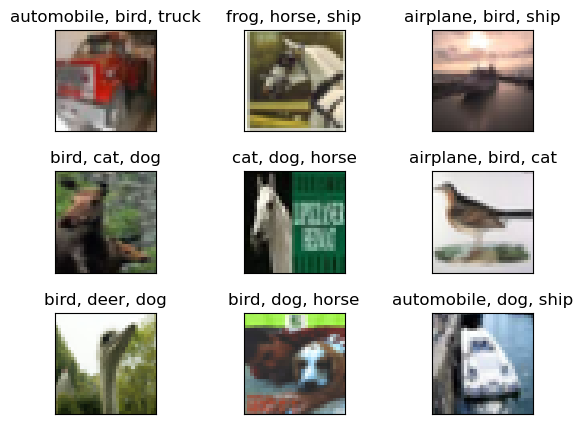}
    \includegraphics[width=0.5\linewidth]{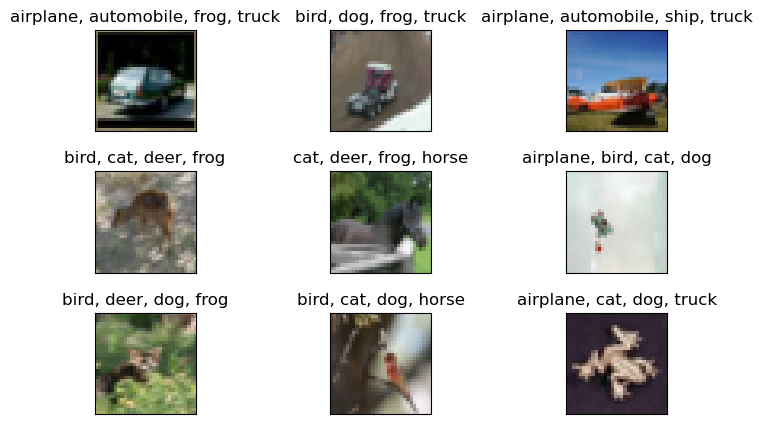}
	\caption{\label{fig:multi} Some images that have multiple labels prediction with their labels.}
\end{figure}

Let us demonstrate  some examples with three and four labels, Figure~\ref{fig:multi}. 

\begin{figure}[H]
	\centering
	\includegraphics[width=0.1\linewidth]{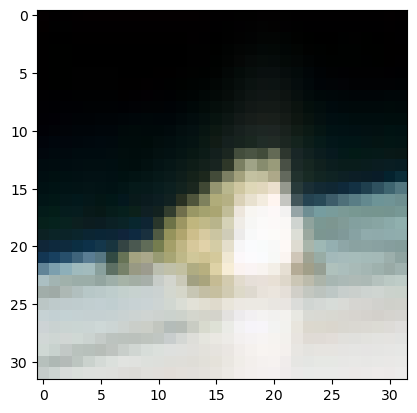}
	\caption{\label{fig:multi5} The image with five labels: 'bird, cat, deer, horse, ship'}
\end{figure}

The only image that has five labels is shown in 
Figure~\ref{fig:multi5}.

\

\textbf{Code Availability} We make our codebase with
model training and conformal predictions available: \cite{AlexAlexD2025}.

\section{Conclusion}

This article has demonstrated a practical approach to leveraging e-conformal prediction with Hoeffding's inequality to enhance the usability of Inductive Conformal Predictors. By employing the concept of e-test statistics and a concentration inequality, we have shown that it is possible to estimate a Hoeffding correction that allows for the repeated use of a single calibration set with a high probability of maintaining the desired marginal coverage.

Our case study on the CIFAR-10 dataset illustrated this methodology. We trained a simple deep neural network and then utilised a calibration set to estimate the empirical mean of the non-conformity scores. By applying Hoeffding's inequality, we determined a correction term $t$ that enabled us to establish a probabilistic bound on the true expectation of the non-conformity measure. This bound was then incorporated into a modified Markov's inequality, leading to a practical rule for constructing prediction sets with a quantifiable confidence level.

The results on the {\em ConformalPredictionTestSet} provided insights into the size distribution of the generated label sets, showcasing the ability of the e-conformal predictor to provide informative uncertainty quantification. While the majority of predictions resulted in single-label sets, the method also identified instances where multiple labels were plausible, reflecting the inherent uncertainty in those cases.

The key takeaway from this work is the potential to mitigate a practical limitation of standard ICPs – the need for a new calibration set for each prediction to guarantee marginal coverage. By demonstrating a method to reuse a calibration set with a high probability of preserving performance, we contribute to making conformal prediction a more readily applicable tool in real-world machine learning scenarios. Future work could explore the tightness of the bounds obtained through Hoeffding's inequality in this context and investigate the application of other concentration inequalities or e-risk control techniques to further optimise the efficiency and reliability of reusable calibration sets in conformal prediction.

\end{document}